\documentclass[10pt,twocolumn,letterpaper]{article}
\pdfoutput=1 

\usepackage{cvpr}
\usepackage{times}
\usepackage{epsfig}
\usepackage{graphicx}
\usepackage{amsmath}
\usepackage{amssymb}

\DeclareMathOperator*{\argmin}{arg\,min}

\usepackage{graphicx}
\usepackage{subfig}

\usepackage[pagebackref=true,breaklinks=true,letterpaper=true,colorlinks,bookmarks=false]{hyperref}

 \cvprfinalcopy 

\ifcvprfinal\fi
\begin{document}

\title{Deep Triplet Ranking Networks for One-Shot Recognition}

\author{Meng Ye\\
Computer and Information Sciences\\
Temple University, USA\\
{\tt\small meng.ye@temple.edu}
\and
Yuhong Guo\\
School of Computer Science\\
Carleton University, Canada \\
{\tt\small yuhong.guo@carleton.ca}
}

\maketitle

\begin{abstract}
Despite the breakthroughs achieved by deep learning models in conventional supervised learning scenarios,
their dependence on sufficient labeled training data in each class prevents effective applications of 
these deep models in situations where labeled training instances for a subset of novel classes 
are very sparse -- in the extreme case only one instance is available for each class.
To tackle this natural and important challenge, 
one-shot learning, which aims to exploit a set of well labeled
base classes to build classifiers for the new target classes that have only one observed instance per class, 
has recently received increasing attention from the research community. 
In this paper we propose a novel end-to-end deep triplet ranking network to perform one-shot learning. 
The proposed approach learns class universal image embeddings 
on the well labeled base classes under a triplet ranking loss,
such that the instances from new classes can be categorized based on their 
similarity with the one-shot instances in the learned embedding space. 
Moreover, our approach can naturally incorporate the available one-shot instances from the new classes
into the embedding learning process to improve the triplet ranking model.
We conduct experiments on two popular datasets for one-shot learning. The results
show the proposed approach achieves better performance than the state-of-the-art comparison methods.
\end{abstract}

\section{Introduction}

Image recognition problem under conventional supervised learning regime has been thoroughly studied, 
generating very successful models like convolutional neural networks (CNNs), 
including AlexNet~\cite{alexnet}, VGG~\cite{vggnet} and ResNet~\cite{resnet}. 
Deep CNN models can achieve impressive performance on large datasets like ImageNet~\cite{imagenet} and Places~\cite{places}. 
However in order to train a deep CNN model, weeks of time and multiple GPUs are typically needed. 
Some works have then used the pre-trained deep neural networks as off-the-shelf feature extractors~\cite{sharif2014cnn}. 
While this procedure is generally promising, it still requires a large amount of labelled images to 
train the model in the first place. 
This is in contrast to the way a human vision system works: 
A person do not need to see tons of image of an object, but only a few of them, to remember and generalize for recognition 
of the objects in the future.
\begin{figure}[t]
   \caption{Example of a 20-way 1-shot classification. Left: \textit{base} classes with plenty of labelled instances for training. Right top: \textit{novel} classes that are disjoint from the \textit{base} ones. Each class only has one labelled instance as training data. Right bottom: test examples. Each test example is to be classified into one of the 20 characters (novel classes). }
\begin{center}
   \includegraphics[width=0.8\linewidth]{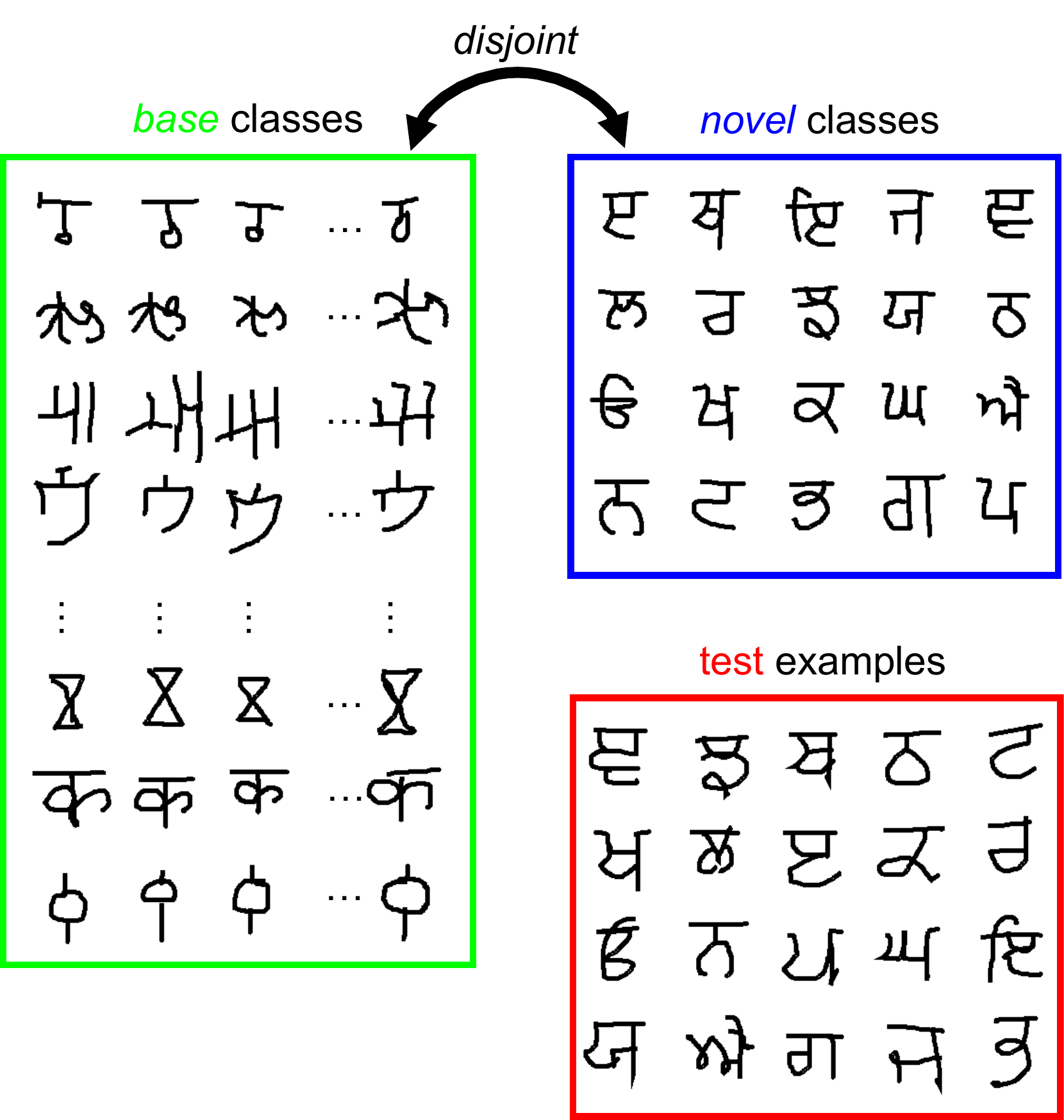}
\end{center}
\label{fig:one-shot}
\end{figure}

Motivated by the gap between the conventional learning and human vision systems, 
one-shot learning and few-shot learning have recently received increasing attention from the research community,
including works on  
one-shot/few-shot learning for character recognition~\cite{koch2015siamese, vinyals2016matching, santoro2016one}, 
image classification~\cite{ravi2016optimization, snell2017prototypical, wang2017multi, wang2016learning, hariharan2016low}, 
face identification~\cite{choe2017face}, image segmentation~\cite{caelles2017one, shaban2017one} 
and seq-to-seq translation\cite{kaiser2017learning}. 
As a counterpart of conventional supervised learning, 
few-shot learning aims to deal with the situation where only a few training instances are available from each class,
while its extreme version, `one-shot learning', tackles the more challenging scenario where
only one instance is available for each class. 
Typically a few-shot learning setting with $N$ novel classes and $k$ instances from each class is referred to as
`N-way, k-shot' learning. 
As learning classifiers solely on one or a few examples from each novel class is extremely difficult, 
studies on one-/few-shot learning have focused on exploiting data from a set of well labeled available base classes.
Figure~\ref{fig:one-shot} presents an example of one-shot learning task in such a setting.

For one-shot learning, due to the extreme sparseness of the training instances in the novel target classes,
a natural learning scheme adopted in the literature is to learn image representations under metric oriented 
learning frameworks that categorize instances based on the similarities between pairs of 
instances~\cite{koch2015siamese}. 
This direct similarity based method however emphasizes more of the absolute similarity values 
during the learning phase without paying substantial attention to the inter-class and intra-class variations,
which can degrade the prediction performance in certain scenarios.
\begin{figure}[t]
   \caption{An example that depicts the strength of relative similarity. \textit{Top}: three characters from three different alphabets. 
The N-ko characters and Armenian characters are very similar in appearance. 
\textit{Middle}: 
A pairwise similarity based model can be confused on a pair of characters from the N-ko and Armenian 
alphabets.
\textit{Bottom}: If a model learns relative similarity from triplet instances, it would not be confused since 
the anchor character (in the red box) is relatively much similar to the N-ko character on the right
than to the Armenian character. 
}
\begin{center}
   \includegraphics[width=0.8\linewidth]{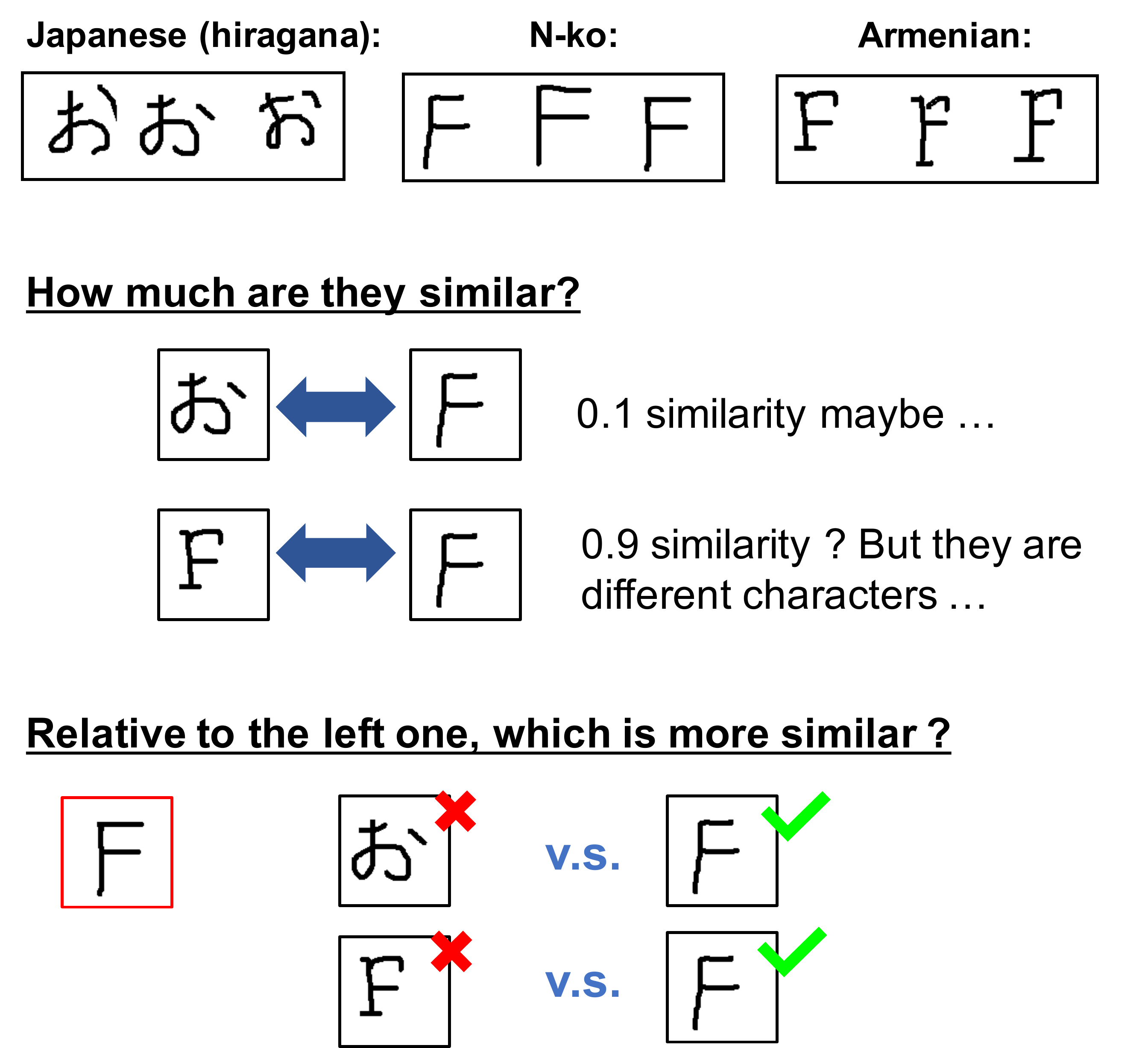}
\end{center}
\label{fig:relative_similarity}
\end{figure}
Hence in this paper we develop a novel image representation based metric learning method
that induce suitable 
\textit{relative similarities} over triplets of images with a deep ranking neural network. 
Our motivation is that a relative similarity based categorization scheme can have larger capacity
in handling inter- and intra-class image variations.
That is when two classes are close in visual appearance or two images from the same class
are far from each other in visual appearance, 
it will be difficult to separate or group them based on absolute similarity measures computed from the visual appearances,
while relative similarities can conveniently handle this 
by only requiring the inter-class similarity to be smaller than intra-class similarity.  
An illustrative example is provided in
Figure~\ref{fig:relative_similarity}. 
Moreover, to increase the model's generalization capacity on the target novel classes, 
we propose to incorporate the one-shot instances from these classes into the deep triplet ranking network
by adding a transformation layer to automatically generate synthetic examples from the one-shot instances.
We conduct experiments on two popular one-shot learning datasets, Omniglot~\cite{lake2015human} and miniImageNet~\cite{ravi2016optimization}. The experimental results demonstrate that the proposed approach achieves the state-of-the-art performance.

\section{Related Work}
\paragraph{One-Shot/Few-Shot Learning} 
Recent works in the literature have adopted different strategies to deal with the few-shot problem in different domains
\cite{koch2015siamese, vinyals2016matching, santoro2016one, ravi2016optimization, snell2017prototypical, wang2017multi, wang2016learning, hariharan2016low, choe2017face, caelles2017one, shaban2017one, kaiser2017learning},
among which the most relevant works are the metric based learning methods 
\cite{koch2015siamese, vinyals2016matching}. 
The authors of \cite{koch2015siamese} proposed a Siamese network to learn direct similarity between image pairs. 
They randomly sample positive and negative pairs and enforce the similarity between the pair of instances
to be either large or small.
As discussed above, this model can induce limited capacity of correctly categorizing difficultly separated 
instances. 
The matching networks developed in \cite{vinyals2016matching} can also be counted as a metric learning approach. 
It minimizes the cosine similarity loss between the one-shot instance and test example,
while embedding the instances  
with a bi-directional LSTM. 
The approach 
still compares pairs in the loss function. 
In the training process, it 
uses `episodes' to update the model, while 
an `episode' contains different set of classes in each time step. 
Another work 
from \cite{ravi2016optimization} also used episodes for training. 
It takes advantage of the periodic training by modelling the learning procedure as a LSTM. 
A more recent work \cite{snell2017prototypical} trained its proposed model on `episodes' for one-shot/few-shot learning as well. 
It proposed to use CNNs as the embedding function and predict class distributions on a given instance 
with a softmax function over the Euclidean distances of the instance to class prototypes.
This model however cannot incorporate the one-shot instances in the learning process.
Another line of research works use networks augmented by external memories for one-shot learning. 
The Memory Augmented Neural Network (MANN) developed in~\cite{santoro2016one} 
enables the information of seen instances to be encoded and retrieved efficiently. 
When exposed to a novel learning task, MANN can rapidly `remember' the new instances and make prediction on test examples. 
The work in \cite{kaiser2017learning} proposed a memory module that can be added to many existing architectures. 
The memory module can be added to Google Neural Machine Translation (GNMT) model to increase performance,
while the memory module augmented CNN achieves the state-of-the-art accuracy on Omniglot dataset~\cite{lake2015human}. 
In addition to these works mentioned above, 
a few other works have 
used data generation to improve one-shot performance \cite{choe2017face, hariharan2016low}.

\vskip .2in
\noindent{\bf Zero-Shot Learning}\quad 
Similar idea of dealing with lack of training data also appears in zero-shot learning. 
As indicated by the name, in zero-shot learning there is not any instance available for the target classes. 
Side information such as label embeddings or attributes is usually used to bridge the gap between seen classes and unseen classes. 
In \cite{lampert2014attribute, romera2015embarrassingly} expert-defined attributes are used to represent animal classes. 
Some common attributes such as `stripe', `spotted' or `live in water' are used to 
serve as an intermediate semantic layer to represent images. 
In \cite{norouzi2013zero, changpinyo2016synthesized} word embeddings trained by Skip-gram models \cite{mikolov2013efficient} 
are used to represent labels in a semantic space  where
similarities between image features and class prototypes can be directly computed. 
In \cite{elhoseiny2013write} the authors used textual descriptions to learn semantic representation of images. 
Then \cite{lei2015predicting} enhanced this idea by using a deep CNN model as an image embedding function. 
Another source of inter-class relation information comes from the web. 
COSTA~\cite{mensink2014costa} uses web image hit-counts to compute co-occurrence statistics as connection between seen and unseen classes for zero-shot learning. In \cite{akata2015evaluation} the authors studied effects of different types of label embeddings on zero-shot performance.

\vskip .2in
\noindent{\bf Metric Learning}\quad 
Another line of relevant research is metric learning. 
There is a large amount of work in the literature of metric learning,
we briefly review a few most relevant works. 
Some early works \cite{chopra2005learning, hadsell2006dimensionality} 
use pairwise constraints to capture similarity between images. 
A few recent works \cite{wang2014learning, schroff2015facenet, hoffer2015deep} adopted CNN structures as embedding functions, 
and use triplet losses instead of pairwise constraints to learn the model. 
Their results demonstrate that combination of deep model and triplet loss is effective in learning similarities. 
Nevertheless the tasks addressed in these metric learning works 
still fall into the conventional learning setting where plenty of labelled data are available. 

\section{Deep Triplet Ranking Networks}
\begin{figure}[t]
   \caption{The framework of the proposed deep triplet ranking network. \textit{Top}: In the training phase, we sample triplets consist of two positive examples and a negative example wrt each class. We augment the data through data 
transformation and then feed them into a CNN. The output embeddings are used to compute the triplet ranking loss, i.e., the optimization objective. 
\textit{Bottom}: In the test phase, both one-shot training instances and test examples are projected to the embedding space. 
Then each test example is assigned to the class of its nearest neighbour among the corresponding one-shot training instances.}
\begin{center}
   \includegraphics[width=0.9\linewidth]{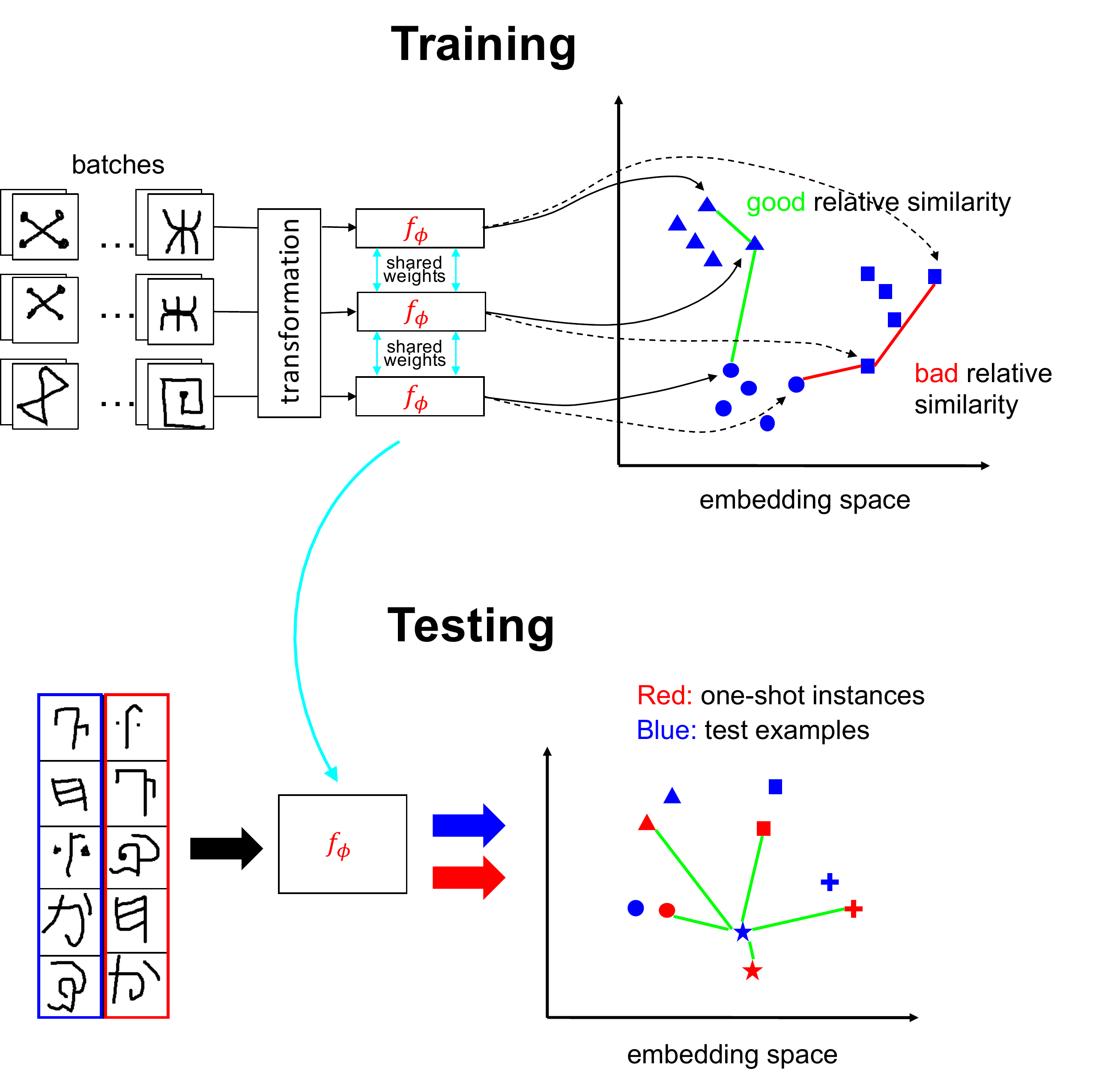}
\end{center}
\label{fig:model}
\end{figure}
\begin{figure*}[t]
   \caption{The CNN architecture constructed for the proposed model. Box with solid lines represent tensors and dashed lines represent different kinds of layers (Red: convolutional layers, Green: max-pooling layers, Blue: fully connected layers). Blocks are separated by max-pooling layers. 
The four blocks contain (2, 2, 3, 3) convolutional layers in each respectively.
Between convolutional layers there are a batch normalization layer and a ReLU non-linearity activation layer. 
After passing through the four blocks, an image will be then mapped into  a 1024-d vector with the final fully connected layer.
}
\begin{center}
   \includegraphics[width=0.9\linewidth]{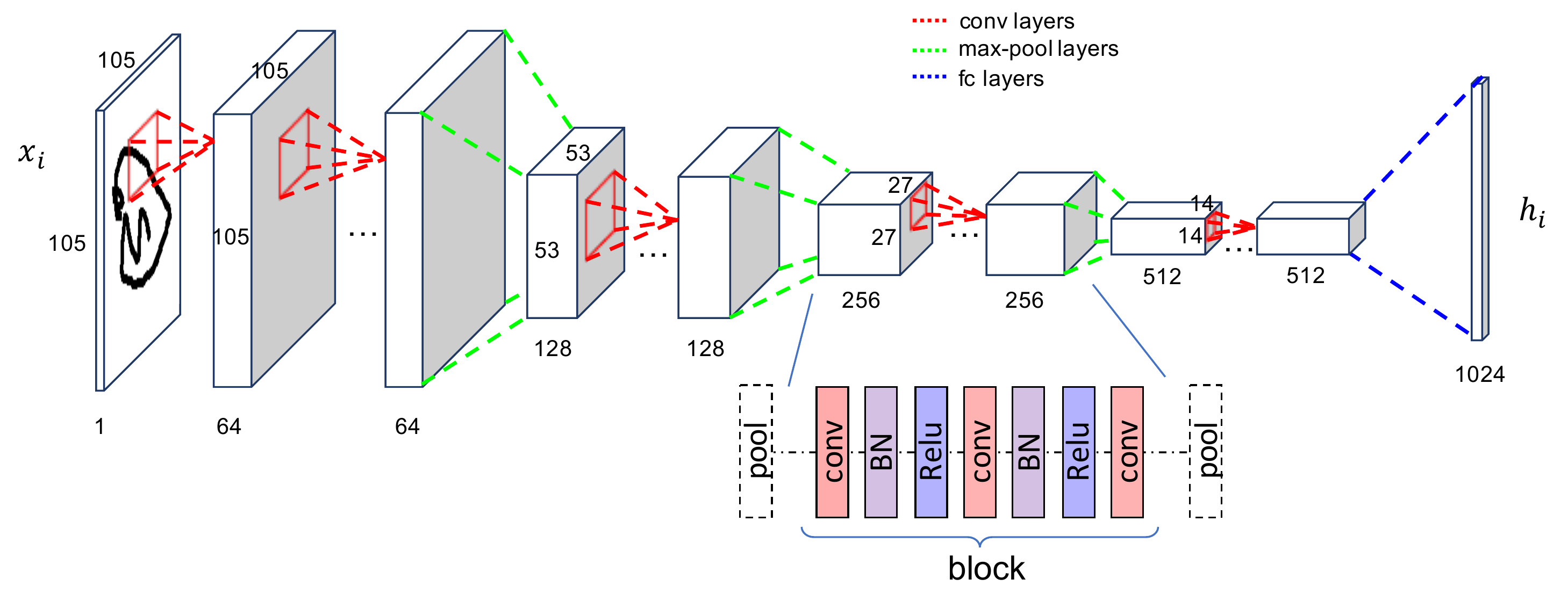}
\end{center}
\label{fig:cnn}
\end{figure*}

In this section we present a novel deep triplet ranking network model for one-shot image recognition. 
We assume there are $N$ labeled training instances from $C$ base classes such that
$\mathcal{D}_{base}=\{(x_i^{base}, y_i^{base})\}_{i=1}^N$ and $y_i^{base}$$\in$$\{B_1,B_2,...,B_C\}$. 
We also assume there is a latent data set from $T$ novel classes such that
$\mathcal{D}_{novel}=\{(x_i^{novel}, y_i^{novel})\}_{i=1}^M$ and $y_i^{novel}$$\in$$\{N_1,N_2,...,N_T\}$,
where only a subset  
$\mathcal{D}_{os}=\{x_k^{o}\}_{k=1}^T$ is sampled as observed one-shot training set. 
The test set $\mathcal{D}_{test}$ which is not observed during training
also comes from the $T$ novel classes.  
The goal of one-shot learning is to train a model 
that can achieve good classification performance on 
the test data $\mathcal{D}_{test}$, 
with only one instance per \textit{novel} class ($\mathcal{D}_{os}$) and plenty of data from the \textit{base} classes ($\mathcal{D}_{base}$). 

The proposed framework is presented in Figure~\ref{fig:model}.
The end-to-end deep training model has two major components. 
First we produce image embeddings with convolutional neural networks (CNNs).
CNNs have demonstrated great power in producing useful image representations in the literature.
We adopted a CNN architecture as our embedding function $f_\phi(\cdot):\mathcal{I}\rightarrow\mathbb{R}^d$, 
where $\mathcal{I}$ denotes the input image space and $\mathbb{R}^d$ denotes the feature embedding space. 
We use $\mathbf{h}$ to denote the output image feature vector such that $\mathbf{h}=f_\phi(x)$. 
Second, given the nature of one-shot learning, we guide the image embedding learning
by using a triplet ranking loss as the training objective. 
With the this triplet ranking loss, 
instances from the same class are moved closer to each other in the embedded space under a given metric
 and those from different classes are pushed apart. 
The one-shot instances can be naturally leveraged in this proposed training framework by 
generating synthetic instances through transformation.
In the test phase, we embed all the one-shot instances in $\mathcal{D}_{os}$ and test examples in $\mathcal{D}_{test}$ 
into the same embedding space and perform a nearest neighbour style comparison to predict the test labels. 

\subsection{Embedding Images with CNNs}
We construct a CNN architecture with four blocks for image embedding.
The architecture is depicted in Figure~\ref{fig:cnn}.
Each block of the architecture contains multiple convolutional layers which extract feature maps at different levels of abstraction. 
As shown in \cite{vggnet}, small size filters can greatly reduce the number of trainable parameters and 
lead to better performance. 
We hence use filters with size of $3\times3$ in each convolutional layer of the CNN architecture.
There are two convolutional layers in each of the first two blocks and three convolutional layers in each of the last two blocks. 
Between every two consecutive convolutional layers in each block, 
a batch normalization (BN) layer \cite{ioffe2015batch} 
and a ReLU non-linearity activation layer are added. 
The batch normalization layer is used to avoid feature scaling problem in batches during training, 
which is usually referred to as `internal covariate shift'. 
We choose ReLU as the activation function because 
it does not suffer from saturation problem and 
can lead to faster training process.

There are 64 filters in each convolutional layer of the first block. 
Then in each consecutive block, it doubles the number of filters 
used in the previous block. 
This leads to 128, 256 and 512 filters in the 2nd, 3rd and 4th block respectively. 
Between every two consecutive blocks, a max-pooling layer with $2\times2$ filter size and stride of 2
is used to spatially down sample the data. 
This typical pooling operation used in many CNN architectures 
can increase model's capacity on dealing with spatial variances and dropping noises from data. 
Meanwhile, the pooling operation halves 
the spatial size of the tensors in the consecutive block. 
An input image of $105\times105$ for the first block 
will reach the size of 
$14\times14$ in the 4th block; 
the output tensor of the last convolutional layer has the size of $14\times14\times512$. 
We finally add a fully-connected (fc) layer with ReLU activation to get a 1024-d vector, 
which is the final embedding feature vector ${\bf h}$ of the input image $x$.
This embedding architecture can be denoted as a mapping function such that $\mathbf{h}=f_\phi(x)$. 
%

\subsection{Triplet Ranking Model}
The key that enables the CNN architecture presented above to 
induce useful image embeddings for performing
image categorization in the novel classes 
is a generalizable learning objective,
which can be optimized to identify the model parameters.
Instead of building prediction models with conventional prediction loss functions,
which are difficult to be generalized into novel classes,
we propose to learn a triplet ranking model 
based on a novel triplet ranking loss.
The triplet ranking loss is based on a relative similarity/distance comparison metric
on triplet instances,
which can be consistent across different classes.
Specifically, 
for any given triplet 
$(x_{pos}^{(1)}, x_{pos}^{(2)}, x_{neg})$, 
where $x_{pos}^{(1)}$ and $x_{pos}^{(2)}$ 
are two distinct instances from same class and $x_{neg}$ comes from a different class as a negative example,
their embedding vectors are produced from the embedding function $f_\phi(\cdot)$ 
such that $\mathbf{h}_{pos}^{(1)}=f_\phi(x_{pos}^{(1)})$, 
$\mathbf{h}_{pos}^{(2)}=f_\phi(x_{pos}^{(2)})$, 
and $\mathbf{h}_{neg}=f_\phi(x_{neg})$. 
The triplet ranking loss is defined as:
\begin{multline}
\mathcal{L}_{triplet}\big(\mathbf{h}_{pos}^{(1)}, \mathbf{h}_{pos}^{(2)}, \mathbf{h}_{neg}\big) = \\
\Big[m + d\big(\mathbf{h}_{pos}^{(1)}, \mathbf{h}_{pos}^{(2)}\big) - d\big(\mathbf{h}_{pos}^{(1)},\mathbf{h}_{neg}\big)\Big]_+ \\
+ \Big[m + d\big(\mathbf{h}_{pos}^{(1)},\mathbf{h}_{pos}^{(2)}\big) - d\big(\mathbf{h}_{pos}^{(2)},\mathbf{h}_{neg}\big)\Big]_+
\end{multline}
where $m$ is a margin which we fix to 2 in the experiments; 
$d(\mathbf{i},\mathbf{j})$ is a metric function that measures distance between vectors $\mathbf{i}$ and $\mathbf{j}$. 
While in our experiments we use a square Euclidean distance $d(\mathbf{i},\mathbf{j})=\|\mathbf{i}-\mathbf{j}\|^2$, 
any other differentiable distance functions can also be used here. 
The capped operator $[x]_+=max(0,x)$ casts the negative part to zero to ensure a ranking hinge loss
that focuses on a relative instead of absolute distance comparison.
This loss function pushes the distance between the two positive examples to be smaller than that
between a positive and a negative examples.  
It guides the training to produce image embeddings that induce metric measures
to achieve this goal.
In practice our model receives data in form of batches during the training phases, 
and the overall loss function is computed over each batch as:
\begin{equation}
\mathcal{L}_{B}=\frac{1}{|B|}\sum_{(\mathbf{p},\mathbf{p'},\mathbf{n})\in B}\mathcal{L}_{triplet}(\mathbf{p}, \mathbf{p'}, \mathbf{n})
\end{equation}
where $B$ is a batch that contains multiple triplets, and the batch size is set to 64 in our experiments. 
The triplets of examples in the batch are sampled independently and uniformly across all the training classes. 

Moreover, with the relative Euclidean distance comparison, the scaling up of the embedding vectors 
can substantially affect the ranking hinge loss -- 
some small distance difference that previously leads to a positive loss 
can become larger than the margin $m$ and thus lead to zero loss. 
We hence propose to add $\ell_2$-norm regularizers over the embedding features  
$\mathbf{h}$ to 
alleviate the impact of the vacuous magnitude increase of ${\bf h}$   
on the learning objective:
\begin{equation}
\mathcal{L}_R = \frac{1}{|B|}\sum_{(\mathbf{p},\mathbf{p'},\mathbf{n})\in B}\big(\|\mathbf{p}\|^2+\|\mathbf{p'}\|^2+\|\mathbf{n}\|^2\big)
\end{equation}
The overall regularized triplet ranking loss is then given by
\begin{equation}
\mathcal{L} = \mathcal{L}_B + \lambda\cdot\mathcal{L}_R
\end{equation}
where $\lambda$ is a trade-off parameter.

After learning a feature embedding function $f_\phi()$ that is universal across classes, 
one-shot prediction is straight forward. For any test example $x_{test}$ from the novel classes, 
we predict its label by finding its nearest neighbour from the one-shot instances
in the embedding space and assigning it to the corresponding class:
\begin{equation}
y^* = \argmin_k \|f_\phi(x_k^{o})-f_\phi(x_{test})\|^2
\end{equation}
If a probability distribution is needed as output, one can simply apply a softmax function: 
\begin{equation}
p(y=k|x_{test})=\frac{exp(-d(f_\phi(x_k^{o}), f_\phi(x_{test})))}{\sum_{k'}exp(-d(f_\phi(x_{k'}^{o}), f_\phi(x_{test})))}
\end{equation}

\subsection{Leveraging One-Shot Instances}
Although the embedding function produced from the proposed deep ranking model
has the generalization capacity over new classes, 
it is a potential information loss to not use
the one-shot instances that directly come from the target classes in the training phase. 
The difficulty of incorporating the one-shot instances under the triplet ranking loss
lies in that there is no anchor point (another positive example) to construct
a triplet over the one-shot example in each target class. 
To solve this problem, we propose to generate synthetic examples to augment the one-shot instances;
in particular,  
for the purpose of simplicity and efficiency 
we use transformation operations to 
produce additional examples from the one-shot instances.
For example, for simple digit datasets, we can use affine transformation operations
such as horizontal shearing, vertical shearing, random rotation, random scaling and random translation 
to generate an augmenting example
$A({\bf x})$ 
for an one-shot image ${\bf x}$. 
For more complex natural images, more complicated transformations that
involve spatial and color adjustments can be used.

We leverage the one-shot instances for model training through a fine tuning procedure. 
After pre-training on the base classes, we proceed to fine tune the model by 
sampling each training batch $B$ in the following way.
For each triplet in the batch, it has half probability to be sampled from the one-shot instances, 
$(x_k^o, A(x_k^o), x_j^o)$, and half probability to be sampled from the base classes, 
$(x_{pos}^{(1)}, x_{pos}^{(2)}, x_{neg})$. 
This gives equal weights to the data from the base classes and the one-shot instances from
the novel target classes.
Whenever the ranking loss on a triplet from the one-shot instances,
$(x_k^o, A(x_k^o), x_j^o)$, 
is positive, it means that two target class prototype embeddings,
$f_\phi(x_k^o)$ and $f_\phi(x_j^o)$, 
are not far apart enough to bear intra-class variance. 
Then the deep network will adjust its weights to 
produce a better embedding function to reduce the loss.
This fine tuning procedure hence is expected to tune the model to produce instance embeddings that 
can better fit the target classes,
while avoiding overfitting the one-shot instances by keeping the base class data.
%

\subsection{Training Algorithm}
We use Adam~\cite{kingma2014adam} to perform stochastic optimization over the learning objective $\mathcal{L}$.
The hyper-parameters of Adam are kept to their default values 
($\beta_1=0.9, \beta_2=0.999$ and $\epsilon=10^{-8}$). 
We set the initial global learning rate as $10^{-4}$, then reduce it by half for every 10k iterations 
to gain a faster convergence. 
Two important issues of deep model training are the weight initialization and overfitting problem. 
Following the literature works, we initialize the weights for each unit of the deep learning model 
with a normal distribution with 
zero mean and standard deviation of $\sqrt{\frac{2}{n}}$, where $n$ is the number of input connections of that unit
\cite{he2015delving}. 
To overcome the overfitting problem, 
we choose to use batch normalization layers between each convolutional layer and ReLU non-linearity layer 
(as depicted in Figure~\ref{fig:cnn}) 
with the Adam algorithm.

\section{Experiment}
To evaluate the effectiveness of the proposed approach, 
we conducted
experiments on two popular one-shot learning datasets, Omniglot~\cite{lake2015human} and miniImageNet~\cite{ravi2016optimization}. 
We present the experimental set up and results in this section.

\subsection{Datasets}
\noindent{\bf Omniglot Dataset}\quad 
This dataset has handwritten characters collected from 50 alphabets. 
The dataset is split into a background set (30 alphabets with 964 characters) and an evaluation set 
(20 alphabets with 659 characters). 
Each character has 20 different images in size of $105\times105\times1$. 
It also provides a test set that consists of 20 sampled subsets from the evaluation set, 
each of which contains 20 characters (classes),
and each class contains two images,  
one for training (one-shot instance) and the other for testing. 
On this dataset, we used affine transformation operations. 
We further split the background set into a training set (20 alphabets with 633 characters) and 
a validation set (10 alphabets with 331 characters) 
and used them to perform hyper-parameter selection. 
After selecting hyper-parameters, 
we adopted the same strategy as in \cite{koch2015siamese}, 
and used the whole background set 
as base classes for training,
while using each subset of the test set as the novel target classes.
The final results reported are the average of 20 runs.
\\

\noindent{\bf miniImageNet Dataset}\quad 
This dataset contains 100 classes randomly sampled from the ImageNet dataset, 
while each class contains 600 images. 
The dataset was further split into a training set of 64 classes, a validation set of 16 and a test set of 20 classes 
\cite{ravi2016optimization}. 
The images have different sizes, and we spatially resized them to the same size of $132\times132\times3$.
Since the images in this dataset contain real objects and natural scenes, 
affine transformations are not suitable as the generated images can have blank borders. 
We transformed the images in the following way. 
First we randomly crop a $105\times105$ region to use. 
Then we randomly flip it horizontally or keep it unchanged, which can model the spatial variance. 
Finally we randomly change the image contrast, which models the colour variance.
Same as in Omniglot, we used training/validation to select hyper-parameters and 
trained the final model on the union of training and validation sets. 
However, on miniImageNet there are no pre-fixed multiple subset samples in the test set. 
Following most literature works, we adopted a '5-way' test setting and
randomly sampled 10 subsets from the test set, each with 5 classes.
Each class contains 1 image as the one-shot instance and 10 images 
as the test examples. 
The results reported are averages over 10 runs, one run for each subset. 
%

\subsection{Comparison Methods}
We compared the proposed approach with three state-of-the-art one-shot learning methods, 
Siamese networks~\cite{koch2015siamese}, Matching networks~\cite{vinyals2016matching} 
and the Prototypical networks~\cite{snell2017prototypical}. 

The Siamese network method developed in \cite{koch2015siamese} is most relevant to 
our proposed approach. Nevertheless, they learn the embedding vectors based on 
pairwise similarities between each pair of instances, while we used relative
distances/similarities over triplets.  
The Matching network model~\cite{vinyals2016matching} 
is a recent work on one-shot learning. They also adopted a metric learning approach. 
However their embedding function for a test image is conditioned on the support set (one-shot instances) 
by using LSTMs. 
The most recent Prototypical network~\cite{snell2017prototypical} is on few-shot learning. 
The authors proposed to represent class prototypes as means of a few examples 
and minimize the distance of instances to their corresponding class prototypes.
This work however cannot exploit the one-shot instances for fine tuning due to the design of their 
training algorithm.

\subsection{Classification Results}
\begin{table}
\caption{20-way 1-shot classification results (\%) on Omniglot. 
Methods marked by \textsuperscript{\textdagger} are implementations from \cite{vinyals2016matching}, 
and methods marked by \textsuperscript{*} are implemented by us.}
\begin{tabular}{l c c} 
 \hline
 \noalign{\vskip 0.1in}
 {\bf Method} & {\bf Fine Tune} & {\bf Accuracy} \\
 \noalign{\vskip 0.1in}
 \hline
 \noalign{\vskip 0.1in}
 Siamese Networks\cite{koch2015siamese} 				& N & 	92.0  \\ 
 Siamese Networks\cite{vinyals2016matching}\textsuperscript{\textdagger} 			& N & 	88.0  \\ 
 Siamese Networks\cite{vinyals2016matching}\textsuperscript{\textdagger}			& Y & 	88.1  \\ 
 Matching Networks\cite{vinyals2016matching} 		& N	&	93.8  \\
 Matching Networks\cite{vinyals2016matching} 		& Y	&	93.5  \\
 Prototypical Networks\cite{snell2017prototypical} 	& N	&	96.0  \\
 \noalign{\vskip 0.1in}
 \hline
 \noalign{\vskip 0.1in}
 Siamese Networks\textsuperscript{*}					& N &	91.5  \\
 Siamese Networks\textsuperscript{*}					& Y &	92.0  \\
 Triplet Ranking Networks 							& N	&	95.5  \\
 Triplet Ranking Networks							& Y	&	{\bf 97.0}  \\
 \noalign{\vskip 0.1in}
 \hline
\end{tabular}
\label{Table:omniglot}
\end{table}

\noindent{\bf One-shot classification on Omniglot}\quad 
We reported the test results of all the comparison methods on the Omniglot dataset 
in Table~\ref{Table:omniglot}. 
Since not every comparison method can exploit the one-shot instances for fine tuning, 
we denoted this specific setup using the `Fine Tune' column
(`N' for not using fine tuning, `Y' for using).
For the Siamese networks, we reported the results from different previous works
\cite{koch2015siamese,vinyals2016matching}.
But in order to obtain a more direct comparison under our CNN architecture with the pairwise loss, 
we also reimplemented the Siamese networks by ourself and conducted experiments. 
We used the same CNN architecture as our proposed model but 
with image pairs as inputs instead of triplets.
We computed the distance of the pair embeddings and passed it through a linear layer 
and a sigmoid function to get a probability value, following the original work \cite{koch2015siamese}. 
For fine tuning, since only one instance is available for each test class, 
we incorporated the pairwise loss where two instance are from different classes. 
From all the comparison results in Table~\ref{Table:omniglot},
we can see our proposed model produces the best result 
by naturally incorporating the one-shot instances.
This suggests that the proposed deep triplet ranking network is effective 
in one-shot learning. 
\\
\begin{table}
\caption{5-way 1-shot classification results (\%) on miniImageNet. 
Methods marked by \textsuperscript{\textdaggerdbl} are implementations from \cite{ravi2016optimization} 
and methods marked by \textsuperscript{*} are implemented by us.}
\begin{tabular}{l c c} 
 \hline
 \noalign{\vskip 0.1in}
 {\bf Method} & {\bf Fine Tune} & {\bf Accuracy} \\
 \noalign{\vskip 0.1in}
 \hline
 \noalign{\vskip 0.1in}
 Baseline 1-NN\cite{ravi2016optimization}\textsuperscript{\textdaggerdbl}
 									& N & 	41.08  \\ 
 Matching Networks\cite{ravi2016optimization}\textsuperscript{\textdaggerdbl}		& N	&	43.40  \\
 Matching Networks FCE\cite{ravi2016optimization}\textsuperscript{\textdaggerdbl}	& N	&	43.56  \\
 Meta-Learner LSTM\cite{ravi2016optimization}\textsuperscript{\textdaggerdbl}		& N &	43.44  \\
 Prototypical Networks\cite{snell2017prototypical} 								& N	&	49.42  \\
 \noalign{\vskip 0.1in}
 \hline
 \noalign{\vskip 0.1in}
 Siamese Networks\textsuperscript{*}					& N &	42.14  \\
 Siamese Networks\textsuperscript{*}					& Y &	43.26  \\
 Triplet Ranking Networks 							& N	&	48.76  \\
 Triplet Ranking Networks							& Y	&	{\bf 50.58}  \\
 \noalign{\vskip 0.1in}
 \hline
\end{tabular}
\label{Table:imagenet}
\end{table}

\noindent{\bf One-shot classification on miniImageNet}\quad 
We then conducted experiments on the miniImageNet dataset.
The images in miniImageNet have more complex object appearances and noises
and are more difficult to tackle than the binary images in Omniglot.
In addition to 
the three comparison methods used above,
on this dataset
we also compared to the results of a Meta-learner LSTM model from \cite{ravi2016optimization}  
and a baseline 1-Nearest Neighbor method.
All the results are reported in Table~\ref{Table:imagenet}. 
We can see that again the proposed triplet ranking network produced the best result
among all the comparison methods. 
Moreover, the direct comparisons with Siamese networks suggest that 
the triplet ranking loss is much more effective than the pairwise loss. 
This again validated the efficacy of the proposed model.\\

\noindent{\bf Ablation study}\quad 
We also conducted experiments on miniImageNet to investigate the impact of different components
of the proposed model, including 
the batch normalization (BN) layers and the embedding regularization term $\mathcal{L}_R$. 
The results are reported in Table~\ref{Table:ablation}. 
By dropping the BN layers from the proposed model, we can see the test accuracy is reduced
from 48.76\% to 46.63\% (without fine tuning). 
This shows the BN layers are important for training the proposed deep model. 
Similarly dropping the embedding regularization term $\mathcal{L}_R$ 
also reduced the test accuracy by about 1\%
(from 48.76\% to 47.79\%).
This validated our assumption on embedding regularization.
\begin{table}
\centering
\caption{Ablation study results (\%) on the proposed model. 
}
\begin{tabular}{l c c} 
\hline
 \noalign{\vskip 0.1in}
 {\bf Method} & {\bf Fine Tune} & {\bf Accuracy} \\
 \noalign{\vskip 0.1in}
 \hline
 \noalign{\vskip 0.1in}
 Proposed - w\textbackslash o BN 				& N	&	46.63  \\
 Proposed - w\textbackslash o $\mathcal{L}_R$	& N	&	47.79  \\
 \noalign{\vskip 0.1in}
 \hline
 \noalign{\vskip 0.1in}
 Proposed 							& N	&	48.76  \\
 Proposed							& Y	&	{\bf 50.58}  \\
 \noalign{\vskip 0.1in}
 \hline
\end{tabular}
\label{Table:ablation}
\end{table}
%

\subsection{Analysis of Deep Image Representations}
In this section we examine the effectiveness of the proposed model
in terms of the image embeddings it produced. 
In order to facilitate one-shot learning, 
the deep embedding function needs to produce image representations that
are generalizable across different classes, in particular to new novel test classes.  
We investigate the representation learning capacity of the proposed deep CNN architecture 
by checking the representations produced by the last fully connected layer 
and the various intermediate convolutional layers along the deep architecture.
\begin{figure*}[t]
\caption{Visualization of the embedding results of different methods on 
instances from 5 different characters (classes) 
in the two dimensional space after 
PCA dimension reduction. 
(a) Siamese Network:  3 similar characters are mixed together on the left. 
(b) Triplet Ranking Network (Proposed):  
Classes are separated much better than the Siamese network. 
But there are still two similar classes mixed on the left bottom corner. 
(c) Triplet Ranking Network with Fine Tuning (Proposed): 
The 5 classes are separated while instances of the same class gathered together. 
}
\subfloat[Siamese Network]
  {\includegraphics[width=.3\linewidth]{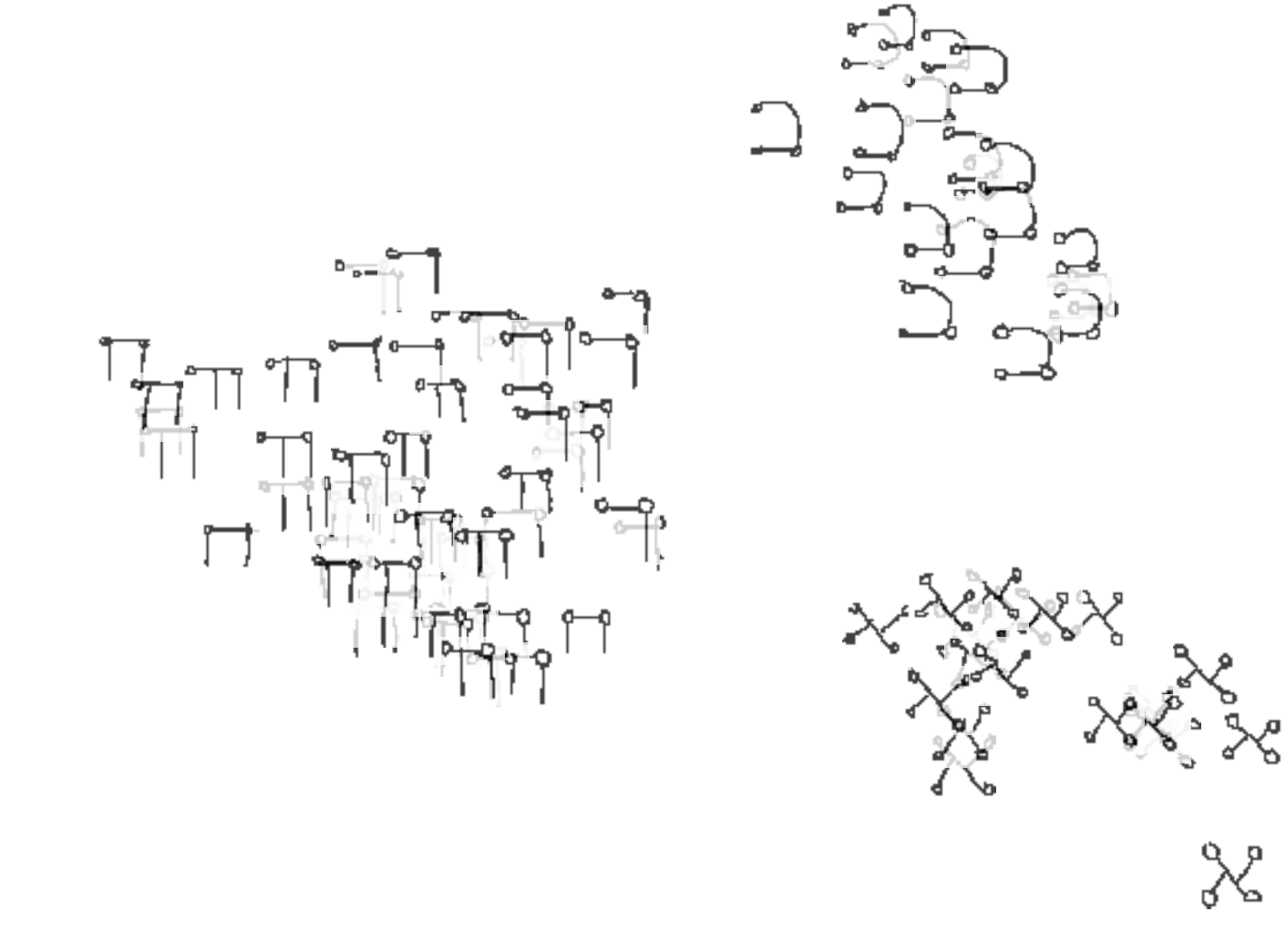}}\hfill
\subfloat[Triplet Ranking Network]
  {\includegraphics[width=.3\linewidth]{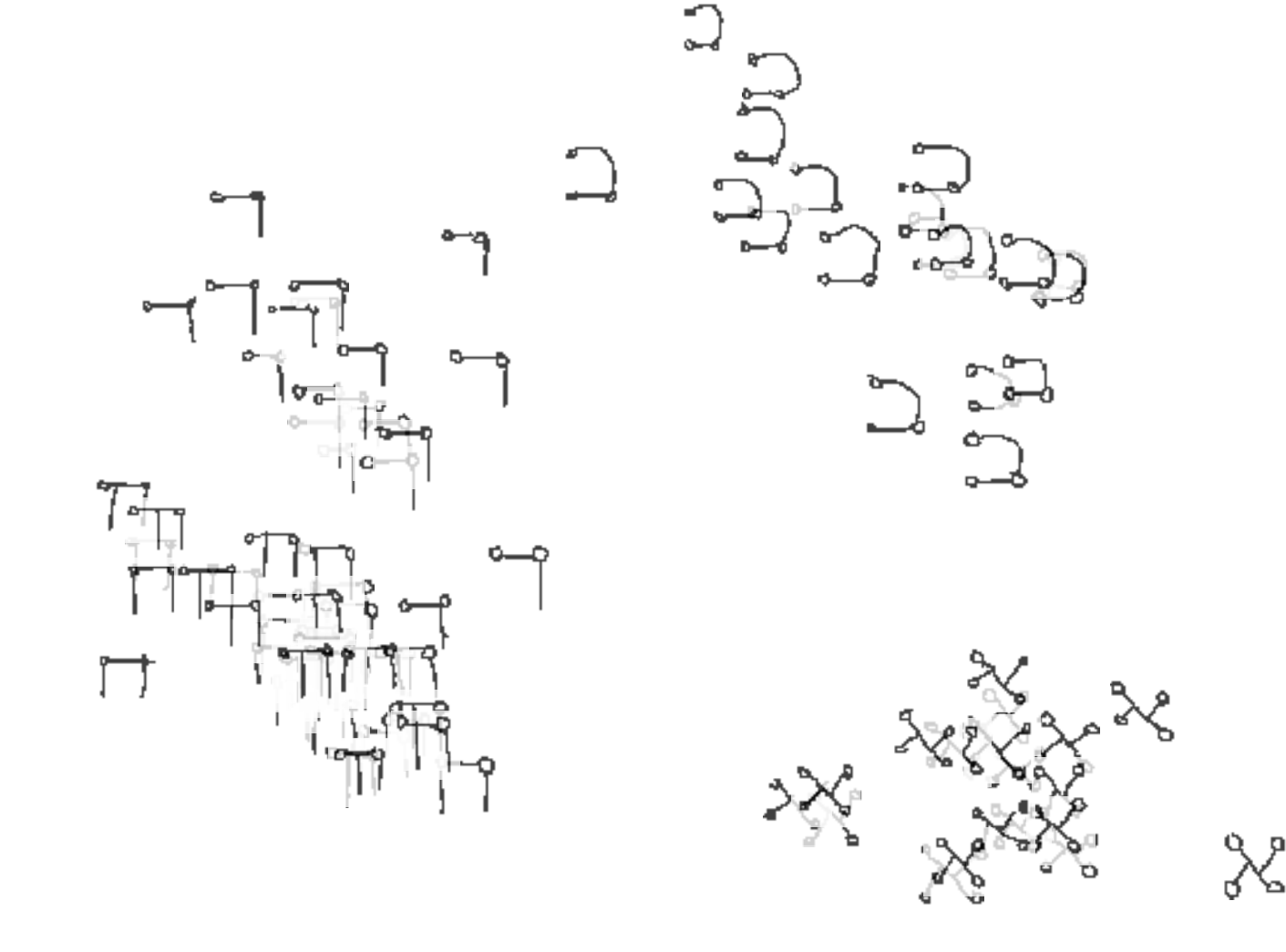}}\hfill
\subfloat[Triplet Ranking Network with Fine Tuning]
  {\includegraphics[width=.3\linewidth]{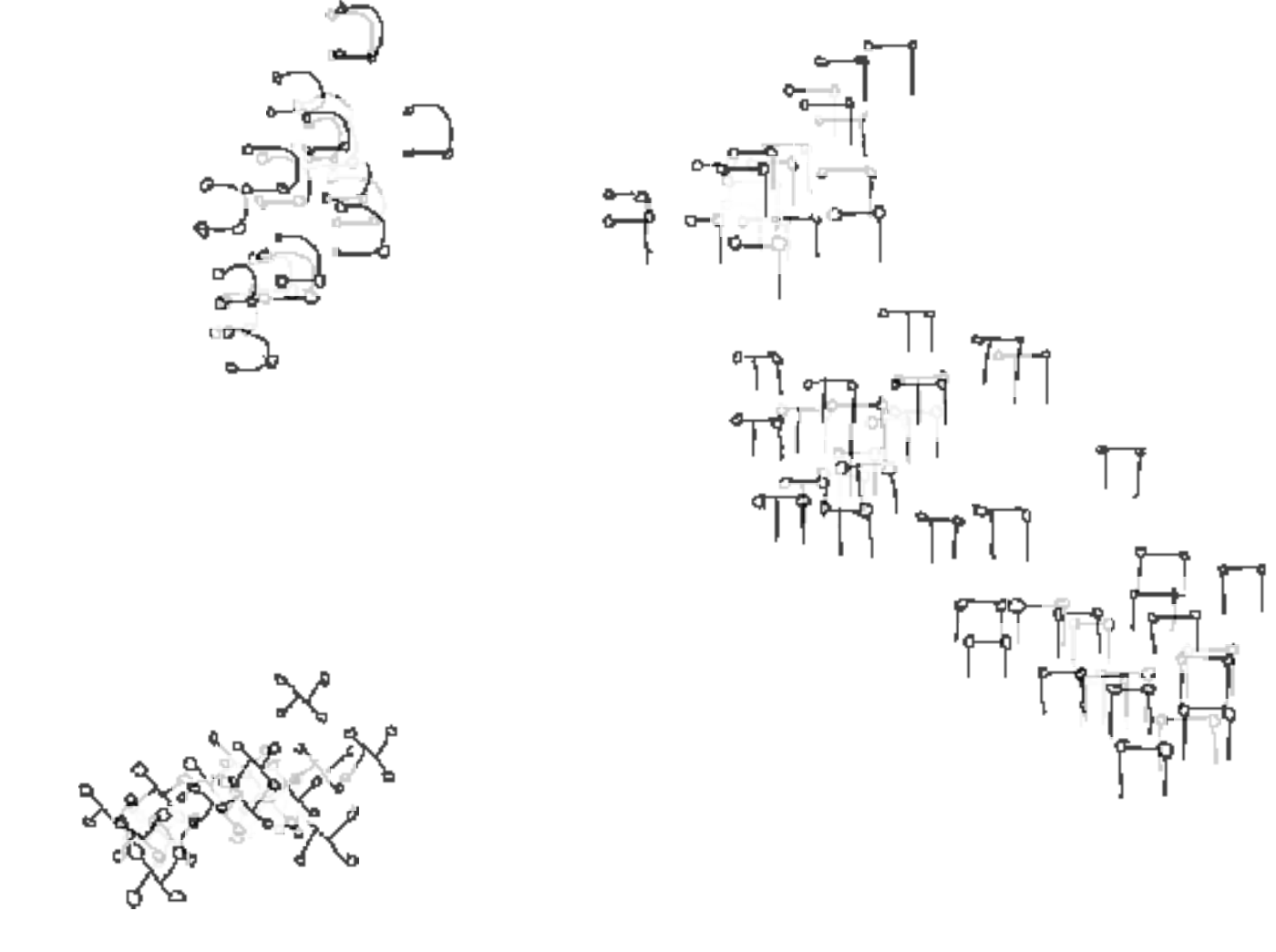}}
\label{fig:embedding}
\vskip -.1in
\end{figure*}

After training, the output vectors of the last fully connected layer for the input images 
are the final image embedding vectors,
which are directly used to compute distances between a test image and the one-shot instances
and perform one-shot classification on the test classes.
Hence the quality of these embeddings are critical to one-shot classification performance. 
To gain a direct and intuitive understanding on the quality of image embeddings produced by the proposed 
deep triplet ranking model, we visualized the embedding vectors of instances from 5 test classes
of the Omniglot dataset in the two dimensional space after dimensionality reduction with PCA. 
The classes are chosen such that some of them are quite similar to each other. 
We compared the embedding results of the proposed deep triplet ranking model with and without 
incorporating the one-shot instances, i.e., fine tuning. 
We also compared with the embedding results of the Siamese networks. 
The visualization results for the three methods are presented in Figure~\ref{fig:embedding}.
We can see that only 3 clusters are recognizable from 
the visualization result of the Siamese network, 
while instances from 3 out of the 5 classes
are mixed together in visualization.   
The visualization result of the deep triplet ranking network without fine tuning is much better
than the Siamese network as one can recognize 4 separated clusters. 
Nevertheless, instances from 2 close classes are still mixed together.  
By naturally incorporating the one-shot instances into the training process, 
i.e., performing fine tuning, the embedding quality of the proposed deep triplet ranking
network is further improved. 
In the visualization result shown in Figure~\ref{fig:embedding} (c),
we can see that 
though the 3 similar classes on the right side still appear close to each other, 
the instances are naturally clustered 
and one can clearly recognize all the 5 clusters.
These results intuitively validated the effectiveness of the proposed model
on producing generalizable image embeddings for one-shot learning.

The deep architecture we constructed for the proposed model 
has multiple convolutional layers. 
Hence the trained model not only can produce the final representation vector
from the fully connected layer, but also can produce intermediate representations
at different level of abstractions
from the convolutional layers. 
We investigated using the intermediate representations of data produced from 
each convolutional layer to perform one-shot classification on Omniglot. 
To get the features from each convolutional layer, we obtain the output tensor from 
the convolutional layer and perform max pooling over the spatial dimensions, 
which resulting in a 1-d feature vector 
with length equal to the number of channels. 
We summarized the average results over 20 runs in Table~\ref{Table:features}. 
We can see that in general higher level layers can produce more effective representations,
which is consistent with the principle of deep learning. 
\begin{table}
\centering
\caption{Average test accuracy (\%) 
on Omniglot
with different levels of feature representations.
}
\begin{tabular}{c c  | c c } 
\hline
 \noalign{\vskip 0.1in}
 {\bf Layer} & {\bf Accuracy} & {\bf Layer} & {\bf Accuracy} \\
 \noalign{\vskip 0.1in}
 \hline
 \noalign{\vskip 0.1in}
 conv-1-1	& 18.3  & conv-3-3	& 57.5  \\
 conv-1-2	& 19.5  & conv-4-1	& 59.8  \\
 conv-2-1	& 22.5  & conv-4-2	& 66.5  \\
 conv-2-2	& 35.0  & conv-4-3	& 85.8  \\
 conv-3-1	& 50.3  & fc-1		& 97.0  \\
 conv-3-2	& 52.0  \\

 \noalign{\vskip 0.1in}
 \hline
\end{tabular}
\label{Table:features}
\vskip -.1in
\end{table}
%

\section{Conclusion}
In this paper we proposed a deep triplet ranking network for one-shot image classification. 
This deep network learns cross-class universal image embeddings 
by optimizing a triplet ranking loss function.
The loss function 
separates the instance pair that belong to the same class from the instance pair
that belong to different classes in the relative distance metric space computed from the image embeddings.
The proposed model can also naturally incorporate one-shot instances into the training process by
generating augmenting instances via image transformations. 
Experiments were conducted on two popularly used one-shot learning datasets.
The results show the proposed model can outperform the state-of-the-art comparison methods.

{\small
\bibliographystyle{ieee}
\bibliography{paperbib}

\begin{thebibliography}{10}\itemsep=-1pt

\bibitem{akata2015evaluation}
Z.~Akata, S.~Reed, D.~Walter, H.~Lee, and B.~Schiele.
\newblock Evaluation of output embeddings for fine-grained image
  classification.
\newblock In {\em CVPR}, 2015.

\bibitem{caelles2017one}
S.~Caelles, K.-K. Maninis, J.~Pont-Tuset, L.~Leal-Taix{\'e}, D.~Cremers, and
  L.~Van~Gool.
\newblock One-shot video object segmentation.
\newblock In {\em CVPR}, 2017.

\bibitem{changpinyo2016synthesized}
S.~Changpinyo, W.-L. Chao, B.~Gong, and F.~Sha.
\newblock Synthesized classifiers for zero-shot learning.
\newblock In {\em CVPR}, 2016.

\bibitem{choe2017face}
J.~Choe, S.~Park, K.~Kim, J.~Hyun~Park, D.~Kim, and H.~Shim.
\newblock Face generation for low-shot learning using generative adversarial
  networks.
\newblock In {\em CVPR}, 2017.

\bibitem{chopra2005learning}
S.~Chopra, R.~Hadsell, and Y.~LeCun.
\newblock Learning a similarity metric discriminatively, with application to
  face verification.
\newblock In {\em CVPR}, 2005.

\bibitem{elhoseiny2013write}
M.~Elhoseiny, B.~Saleh, and A.~Elgammal.
\newblock Write a classifier: Zero-shot learning using purely textual
  descriptions.
\newblock In {\em ICCV}, 2013.

\bibitem{hadsell2006dimensionality}
R.~Hadsell, S.~Chopra, and Y.~LeCun.
\newblock Dimensionality reduction by learning an invariant mapping.
\newblock In {\em CVPR}, 2006.

\bibitem{hariharan2016low}
B.~Hariharan and R.~Girshick.
\newblock Low-shot visual recognition by shrinking and hallucinating features.
\newblock {\em arXiv preprint arXiv:1606.02819}, 2016.

\bibitem{he2015delving}
K.~He, X.~Zhang, S.~Ren, and J.~Sun.
\newblock Delving deep into rectifiers: Surpassing human-level performance on
  imagenet classification.
\newblock In {\em ICCV}, 2015.

\bibitem{resnet}
K.~He, X.~Zhang, S.~Ren, and J.~Sun.
\newblock Deep residual learning for image recognition.
\newblock In {\em CVPR}, 2016.

\bibitem{hoffer2015deep}
E.~Hoffer and N.~Ailon.
\newblock Deep metric learning using triplet network.
\newblock In {\em International Workshop on Similarity-Based Pattern
  Recognition}, 2015.

\bibitem{ioffe2015batch}
S.~Ioffe and C.~Szegedy.
\newblock Batch normalization: Accelerating deep network training by reducing
  internal covariate shift.
\newblock In {\em ICML}, 2015.

\bibitem{kaiser2017learning}
{\L}.~Kaiser, O.~Nachum, A.~Roy, and S.~Bengio.
\newblock Learning to remember rare events.
\newblock {\em arXiv preprint arXiv:1703.03129}, 2017.

\bibitem{kingma2014adam}
D.~Kingma and J.~Ba.
\newblock Adam: A method for stochastic optimization.
\newblock {\em arXiv preprint arXiv:1412.6980}, 2014.

\bibitem{koch2015siamese}
G.~Koch, R.~Zemel, and R.~Salakhutdinov.
\newblock Siamese neural networks for one-shot image recognition.
\newblock In {\em ICML Deep Learning Workshop}, 2015.

\bibitem{alexnet}
A.~Krizhevsky, I.~Sutskever, and G.~E. Hinton.
\newblock Imagenet classification with deep convolutional neural networks.
\newblock In {\em NIPS}, 2012.

\bibitem{lake2015human}
B.~M. Lake, R.~Salakhutdinov, and J.~B. Tenenbaum.
\newblock Human-level concept learning through probabilistic program induction.
\newblock {\em Science}, 350(6266):1332--1338, 2015.

\bibitem{lampert2014attribute}
C.~H. Lampert, H.~Nickisch, and S.~Harmeling.
\newblock Attribute-based classification for zero-shot visual object
  categorization.
\newblock {\em TPAMI}, 36(3):453--465, 2014.

\bibitem{lei2015predicting}
J.~Lei~Ba, K.~Swersky, S.~Fidler, et~al.
\newblock Predicting deep zero-shot convolutional neural networks using textual
  descriptions.
\newblock In {\em ICCV}, 2015.

\bibitem{mensink2014costa}
T.~Mensink, E.~Gavves, and C.~G. Snoek.
\newblock Costa: Co-occurrence statistics for zero-shot classification.
\newblock In {\em CVPR}, 2014.

\bibitem{mikolov2013efficient}
T.~Mikolov, K.~Chen, G.~Corrado, and J.~Dean.
\newblock Efficient estimation of word representations in vector space.
\newblock In {\em ICLR}, 2013.

\bibitem{norouzi2013zero}
M.~Norouzi, T.~Mikolov, S.~Bengio, Y.~Singer, J.~Shlens, A.~Frome, G.~S.
  Corrado, and J.~Dean.
\newblock Zero-shot learning by convex combination of semantic embeddings.
\newblock {\em arXiv preprint arXiv:1312.5650}, 2013.

\bibitem{ravi2016optimization}
S.~Ravi and H.~Larochelle.
\newblock Optimization as a model for few-shot learning.
\newblock In {\em ICLR}, 2017.

\bibitem{romera2015embarrassingly}
B.~Romera-Paredes and P.~Torr.
\newblock An embarrassingly simple approach to zero-shot learning.
\newblock In {\em ICML}, 2015.

\bibitem{imagenet}
O.~Russakovsky, J.~Deng, H.~Su, J.~Krause, S.~Satheesh, S.~Ma, Z.~Huang,
  A.~Karpathy, A.~Khosla, M.~Bernstein, A.~C. Berg, and L.~Fei-Fei.
\newblock {ImageNet Large Scale Visual Recognition Challenge}.
\newblock {\em IJCV}, 115(3):211--252, 2015.

\bibitem{santoro2016one}
A.~Santoro, S.~Bartunov, M.~Botvinick, D.~Wierstra, and T.~Lillicrap.
\newblock One-shot learning with memory-augmented neural networks.
\newblock {\em arXiv preprint arXiv:1605.06065}, 2016.

\bibitem{schroff2015facenet}
F.~Schroff, D.~Kalenichenko, and J.~Philbin.
\newblock Facenet: A unified embedding for face recognition and clustering.
\newblock In {\em CVPR}, 2015.

\bibitem{shaban2017one}
A.~Shaban, S.~Bansal, Z.~Liu, I.~Essa, and B.~Boots.
\newblock One-shot learning for semantic segmentation.
\newblock {\em arXiv preprint arXiv:1709.03410}, 2017.

\bibitem{sharif2014cnn}
A.~Sharif~Razavian, H.~Azizpour, J.~Sullivan, and S.~Carlsson.
\newblock Cnn features off-the-shelf: an astounding baseline for recognition.
\newblock In {\em CVPR workshops}, 2014.

\bibitem{vggnet}
K.~Simonyan and A.~Zisserman.
\newblock Very deep convolutional networks for large-scale image recognition.
\newblock {\em arXiv preprint arXiv:1409.1556}, 2014.

\bibitem{snell2017prototypical}
J.~Snell, K.~Swersky, and R.~S. Zemel.
\newblock Prototypical networks for few-shot learning.
\newblock {\em arXiv preprint arXiv:1703.05175}, 2017.

\bibitem{vinyals2016matching}
O.~Vinyals, C.~Blundell, T.~Lillicrap, D.~Wierstra, et~al.
\newblock Matching networks for one shot learning.
\newblock In {\em NIPS}, 2016.

\bibitem{wang2014learning}
J.~Wang, Y.~Song, T.~Leung, C.~Rosenberg, J.~Wang, J.~Philbin, B.~Chen, and
  Y.~Wu.
\newblock Learning fine-grained image similarity with deep ranking.
\newblock In {\em CVPR}, 2014.

\bibitem{wang2017multi}
P.~Wang, L.~Liu, C.~Shen, Z.~Huang, A.~van~den Hengel, and H.~T. Shen.
\newblock Multi-attention network for one shot learning.
\newblock In {\em CVPR}, 2017.

\bibitem{wang2016learning}
Y.-X. Wang and M.~Hebert.
\newblock Learning to learn: Model regression networks for easy small sample
  learning.
\newblock In {\em ECCV}, 2016.

\bibitem{places}
B.~Zhou, A.~Lapedriza, J.~Xiao, A.~Torralba, and A.~Oliva.
\newblock Learning deep features for scene recognition using places database.
\newblock In {\em NIPS}, 2014.

\end{thebibliography}
}

\end{document}